# A Two-Phase Safe Vehicle Routing and Scheduling Problem: Formulations and Solution Algorithms

Aschkan Omidvar [a*], Eren Erman Ozguven [b], O. Arda Vanli [c],
R. Tavakkoli-Moghaddam [d]

[a] *Department of Civil and Coastal Engineering, University of Florida, Gainesville, FL 32611, USA*
[b] *Department of Civil and Environmental Engineering, FAMU-FSU College of Engineering., Tallahassee, FL, 32310, USA*
[c] *Department of Industrial and Manufacturing Engineering, FAMU-FSU College of Engineering, Tallahassee, FL, 32310, USA*
[d] *School of Industrial Engineering, College of Engineering, University of Tehran, Iran*

---

**Abstract**

We propose a two phase time dependent vehicle routing and scheduling optimization model that identifies the safest routes, as a substitute for the classical objectives given in the literature such as shortest distance or travel time, through (1) avoiding recurring congestions, and (2) selecting routes that have a lower probability of crash occurrences and non-recurring congestion caused by those crashes. In the first phase, we solve a mixed-integer programming model which takes the dynamic speed variations into account on a graph of roadway networks according to the time of day, and identify the routing of a fleet and sequence of nodes on the safest feasible paths. Second phase considers each route as an independent transit path (fixed route with fixed node sequences), and tries to avoid congestion by rescheduling the departure times of each vehicle from each node, and by adjusting the sub-optimal speed on each arc. A modified simulated annealing (SA) algorithm is formulated to solve both complex models iteratively, which is found to be capable of providing solutions in a considerably short amount of time. In this paper, speed (and travel time) variation with respect to the hour of the day is calculated via queuing models (i.e., $M/G/1$) to capture the stochasticity of travel times more accurately unlike the most researches in this area, which assume the speed on arcs to be a fixed value or a time dependent step function. First, we demonstrate the accurate performance of $M/G/1$ in estimation and predicting speeds and travel times for those arcs without readily available speed data. Crash data, on the other hand, is obtained for each arc. Next, 24 scenarios, which correspond to each hour of a day, are developed, and are fed to the proposed solution algorithms. This is followed by evaluating the routing schema for each scenario where the following objective functions are utilized: (1) the minimization of the traffic delay (maximum congestion avoidance), and (2) the minimization of the traffic crash risk, and (3) the combination of two objectives. Using these objectives, we identify the safest routes, as a substitute for the classical objectives given in the literature such as shortest distance or travel time, through (1) avoiding recurring congestions, and (2) selecting routes that have a lower probability of crash occurrences and non-recurring congestion caused by those crashes. This also allows us to discuss the feasibility and applicability of our model. Finally, the proposed methodology is applied on a benchmark network as well as a small real-world case study application for the City of Miami, Florida. Results suggest that in some instances, both the travelled distance and travel time increase in return for a safer route, however, the advantages of safer route can outweigh this slight increase.

---

* Corresponding author. Tel.: +1-850-405-6688
E-mail address: Aschkan@ufl.edu

---

## 1. Introduction and literature reivew

Traffic crashes and congestion are major costs to the collective and social well-being. According to the Federal Highway Administration, traffic crashes imposed an economic cost of $242.0 billion to the U.S. economy in 2010 [1]. Total estimated cost of congestion to Americans was also approximately $124 billion in 2013 [1]. These figures show the vital influence of congestion and crashes on our daily trips. For many years, vehicle routing and scheduling have been used to investigate the effects of congestion on the roadway networks. There are two main sources of congestion: recurring and non-recurring. The literature shows that in the United States, 40% of traffic congestion is recurring due to spatial, traffic and behavioural issues, and traffic accidents, construction, work zones and environmental events contribute to 25%, 15% and 10% of the total traffic congestion, respectively [1]. Without a doubt, travel safety is one of the main components of transportation activities, however, the idea of safety in the field of vehicle routing is not as maturely developed as transportation planning or logistics. Hence, re-routing vehicle through safer road segments and intersections can be one of many strategies to increase safety in micro and macro scale. An optimization approach to minimize route hazard elements could be a handy mean to achieve safer routes.

The origins of fleet routing problem goes as early as Dantzig and Ramser [2], who proposed the first vehicle routing problem (VRP) in the context of truck dispatching for gas delivery to fuel stations. This paper opened a new chapter in combinatorial optimization, and numerous researchers applied this for other problems such as waste collection, school bus routing, routing in supply chain, dial-a-ride services, and other goods/services collection and dispatching problems. Eventually, Clarke and Wright [3] improved this model, and proposed a heuristic algorithm to solve the VRP problem. The basic VRP consists a set of homogeneous vehicles with limited capacity and a central depot. Vehicles serve the customers located in each node of the graph through the arcs between each pair of nodes. Each customer has a specific demand, shown as $q_i$ and travelling on each arc has a linear correlation with the cost components. All vehicles must return to the central depot after serving all the customers [4]. However, these conditions may change depending on the type of VRP, such as pick-up and delivery VRP, capacitated VRP, and open VRP among others.

Vehicle Routing and Scheduling Problem involves the activities of route assignment for dispatching and/or collecting from/to one or multiple depot to/from customer nodes, which are scattered geographically. These problems usually come with several constraints, such as vehicle capacity, maximum travelled distance or time, time windows or customers prioritizing. VRP is defined on a directed or undirected graph as $G(V, A)$, where $V = (0, 1, \dots, n)$ is the vector of nodes and $A = ((i, j): i, j \epsilon V, i \neq j)$ represents the set of arcs. Each arc contains a positive cost of $c_{ij}$, which can be distance, time, or any other cost component. The goal is to find one or several Hamiltonian loops, where all customers are being served with the minimum possible cost.

In most VRPs, the speeds and travel times on arcs are assumed to be constant. In other words, Euclidean distances are assumed with a fixed cost function on each arc. On the other hand, stochastic VRP has mainly focused on nodes than arcs. In other words, these models study demand or service time at nodes, and rarely studied speed and other travel characteristics on arcs. By contrast, dynamic VRP has considered changing travel times; however, technology, such as wireless systems, GPS, short signals among others, is a must in this type of VRP, and routing and planning is infeasible without this technology. Fixed travel cost function in today's volatile and uncertain transportation is not feasible. Therefore, time-dependent VRP (TDVRP) was first introduced by Cooke and Halsey (1966) where they expanded the idea of shortest path between nodes in a network from fixed times to variable times. However, they did not consider the case of multiple vehicles. In the TDVRP, the travel cost function is assumed to be changing according to the time of the day [5]. This assumption enables us to consider the speed variations on the arcs, which can help a roadway user avoid congestion [6, 7]. This extension of VRP was initially proposed at the early stages of research on VRP. However, due to its complexity in modelling and solution, it has not been studied until the past decade. Malandraki and Dastkin (1992) proposed heuristic neighborhood search algorithms to solve the TDVRP and time-dependent travelling salesman problem (TDTSP), where they suggested a step function in order to consider travel time variations throughout the planning horizon.

Most researchers considered travel times at discrete levels (Table 1. Research on Time-dependent Vehicle Routing Problem (TDVRP). Although a step function for travel times and speeds can simulate the real world condition better than a fixed travel time (or speed) value, there is a need for further development to successfully represent the actual real-life speed variations and changes in the traffic flow on roadways [8, 9]. Other approaches to incorporate parameter variations include dynamic and stochastic vehicle routing applications that focus on the service time and demand at the nodes [10], models considering speed variations and dynamic travel times [11-13], and stochastic and time-varying network applications [14, 15]. For further information on the static and dynamic vehicle routing problems (VRP), please refer to [4, 10].

Table 1. Research on Time-dependent Vehicle Routing Problem (TDVRP)

| Authors | Model Type | Research Description | Solution Approach |
|---|---|---|---|
| [16] | Basic | Travel time variation is ignored. Vehicle load and traffic congestion in peak hours are considered. | Integer Programming |
| [11] | Discrete travel times and objective function | MILP with discrete step function for travel times. | Branch & cut and Greedy neighborhood search |
| [17] | Discrete travel times and objective function | Modelling based on travel times according to the dispatching time | All alternative solutions on small size case studies up to 5 nodes are studied |
| [5] | Discrete travel times and objective function | Modelling based on travel times according to the dispatching time | Tabu search on Solomon benchmark instances |
| [18] | Discrete travel times and objective function | Modelling based on travel times according to the dispatching time | Local Search |
| [19] | Discrete travel times and objective function | Modelling based on travel times according to the dispatching time | Ant Colony Optimization |

To the best of authors' knowledge, traffic safety, in terms of crash risk on roadways, has not been introduced to the graph theory and transportation network optimization. Therefore, this study is an important step towards filling this gap. In this paper, we propose a two phase vehicle routing and scheduling optimization model that identifies the safest routes, as a substitute for the classical objectives given in the literature such as shortest distance or travel time, through (1) avoiding recurring congestions, and (2) selecting routes that have a lower probability of crash occurrences and non-recurring congestion caused by those crashes.

## 2. Mathematical Modelling

The proposed modeling approach has two phases as seen in Fig. 1. We will discuss these phases in the following subsections.

### *2.1. Phase 1:Routing graph modelling*

In this phase, we formulate a mixed-integer programming which takes the dynamic speed variations into account on a graph of roadway networks, according to the time of day. The probability of crash as a function of the speed is calculated based on traffic crash record for three years on each roadway segment and intersection. In most locations, with an increase in the traffic density, and reduction in speed, the probability of having a crash increases [20-22].

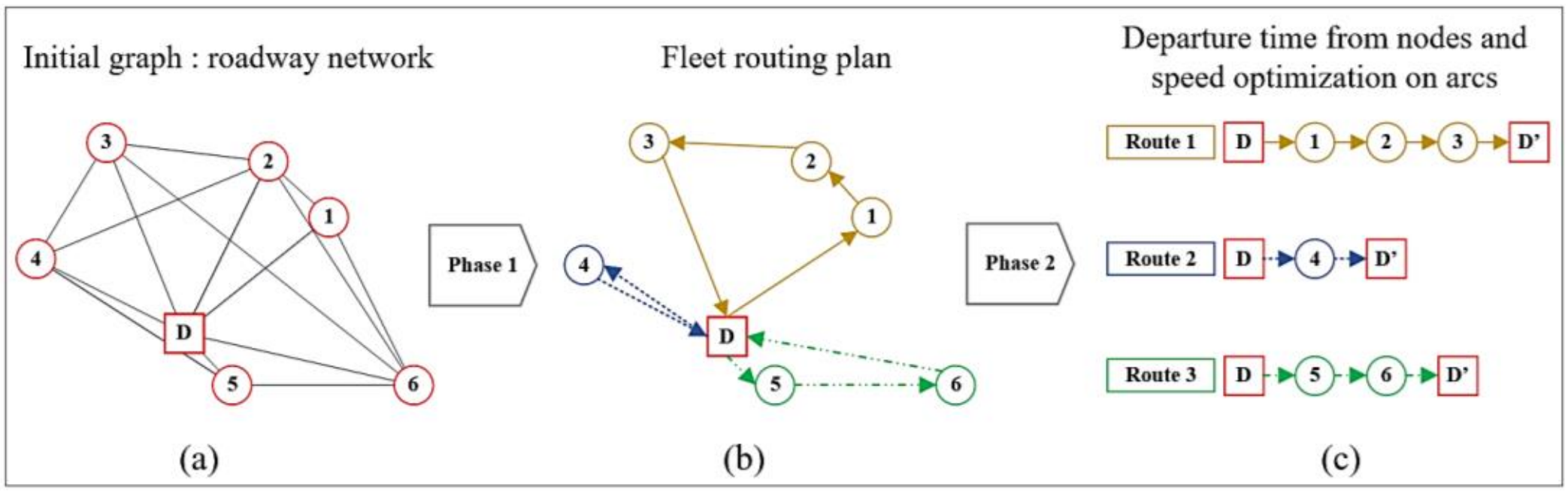

Fig. 1. A schematic representation of modelling approach: (a) Roadway network (b) Result of Phase 1: routes identified (c) Result of Phase 2: route schedules determined, including departure time and the travel speed at each node

Therefore, the first graph model identifies the routing of a fleet and sequence of nodes on the safest feasible paths. Several constraints, such as hard and soft time windows on each node, capacity, operation hours, and number of vehicles are introduced to ensure the fast and quality service. The model takes real time speed data as input. However, in case the user only has access to the traffic flow data but not speed data, we also propose a methodology to incorporate the speed variation with respect to the hour of the day, which is obtained based on (a) the available traffic flow values, and (b) the queuing model concept (namely *M/G/1*) to capture the stochasticity of speed variations and travel times more accurately. Please see Section 2.3 for this methodology. So, in our first model, travel times are changeable and they are a function of speed. The objective function consists of two main components: (1) crash probability on each segment according to the time of the day and (2) normalized Travel Time Index (TTI) [23], which is a function of travel time. In summary, the two components of the objective function allows one to (a) increase the trip safety by choosing segments with lower crash risk, and (b) avoid recurring and non-recurring congested segments. Fig. 1a shows the initial network, and Fig. 1b indicates the routes identified as a result of Phase 1.

The first model is defined on a Hamiltonian graph as $G$= ($V$, $A$) for which $V$= $(v_0, \dots, v_{n+1})$ represent the set of nodes. Nodes $v_0$ and $v_{n+1}$ refer to the central depot, and the nodes $(v_1, \dots, v_n)$ represent a set of customer nodes. $A$= $((i, j)$:$i, j \in V, i \neq j)$ is the set of arcs. $K$ demonstrates the number of fleets at the time of start ready in the depot, $K$= $(1,\dots,k)$. Moreover, $q_i$ represents the non-negative demand at node $i$ and the maximum load for each vehicle is denoted as $Q_k$. $s_i$ stands for non-negative service time at the node $i$. Service time and demand for the central depot is assumed as zero ($s_0 = 0$, $q_0 = 0$). Customer node has a service time window of $[e_i, l_i]$, and $L$ is the latest allowed time for the departing depot. Each arc is associated with a fixed distance ($d_{ij}$).

We incorporate three time dependent parameters in the model: (1) the departure time from node $i$ ($p_{ik}$), (2) travel time between nodes $i$ and $j$ in hours $t_{ij}$ ($p_{ik}$), which is a function of ($p_{ik}$), and (3) $TT_{ijk}$ ($p_{ik}$) is the time index from $i$ to $j$ for vehicle $k$ when the vehicle leaves the node $i$ at the time $p_{ik}$. Finally, the crash probability for arc $(i,j)$ at the departure time ($p_{ik}$) is shown as $\xi_{ij}$ ($p_{ik}$) which is a non-negative and non-zero value with the maximum of 1.

The first decision variable of the model is $x_{ijk}$, which is a binary variable that takes a value of 1 if vehicle $k$ leaves node $i$ for node $j$, and otherwise takes a value of 0. $w_{ik}$ demonstrates the amount of net load that the vehicle $k$ carries after leaving the node $i$. $a_{ik}$ and $p_{ik}$ track the times that vehicle k enters and leaves node $i$, respectively.

The optimization model for the first phase is presented as follows:

Objectives:

$$minimize \;\; 1-\left[\prod_{k=1}^{K}\prod_{(i,j)\in V}\left[1-\left(\xi_{ijk}(p_{ik})\right)\right]^{x_{ijk}}\right] \tag{1}$$

$$minimize \;\; \sum_{k=1}^{K}\sum_{(i,j)\in V} TT_{ijk}(p_{ik})\; x_{ijk} \tag{2}$$

Subject to:

$$\sum_{k=1}^{K}\sum_{j=0}^{n+1} x_{ijk} = 1 \ ; \ \forall i \in \{1, \dots, n\} \tag{3}$$

$$\sum_{j=1}^{n}\sum_{k=1}^{K} x_{0jk} \ \leq K \tag{4}$$

$$\sum_{j=1}^{n}\sum_{k=1}^{K} x_{0jk} = \sum_{i=1}^{n}\sum_{k=1}^{K} x_{i,n+1,k} \tag{5}$$

$$\sum_{i=0}^{n} x_{ifk} - \sum_{j=1}^{n+1} x_{fjk} = 0 \ ; \ \forall k \in K \ , f \in \{1, \dots, n\} \tag{6}$$

$$x_{i0k} + x_{n+1,i,k} = 0 \ ; \ \forall k \in K \ , \ \forall i \in \{1, \dots, n\} \tag{7}$$

$$\sum_{\substack{i=1 \\ i \neq j}}^{n} q_i \sum_{\substack{j=1 \\ j \neq i}}^{n+1} x_{ijk} \leq Q_k \ ; \ \forall \ k \in K \tag{8}$$

$$e_i \sum_{j=1}^{n+1} x_{ijk} \leq a_{ik} \leq l_i \sum_{j=1}^{n+1} x_{ijk} \ ; \ \forall i \in \{0, 1, \dots, n+1\} \quad , \quad \forall k \in K \tag{9}$$

$$a_{ik} + s_i + t_{ijk}(p_{ik}) \leq a_{jk} \times x_{ijk} + \left(1 - x_{ijk}\right) \times L \ ; \ \forall i, j \in \{0, 1, \dots, n+1\} \ , \ \forall k \in K \tag{10}$$

$$a_{ik} + s_i \leq p_{ik} \leq L - t_{i,n+1} \ ; \ \forall k \in K \ , \ \forall \ i \in \{0,1, \dots, n\} \tag{11}$$

$$a_{ik} \geq 0 \ ; \ \forall i \in \{0,1,\dots,n+1\} \ ; \ \forall k \in K \tag{12}$$

$$x_{ijk} \in \{0,1\} \ ; \ \forall (i,j) \in A \ , \ \forall k \in K \tag{13}$$

$$w_{ik} \geq 0 \ ; \ \forall i \in \{0,1,\dots,n+1\} \ ; \ \forall k \in K \tag{14}$$

Objective function (1) minimizes the total crash probability on all routes. Objective function (2) minimizes the total travel time index (TTI). Travel time index $TT_{ijk}(p_{ik})$ is defined as the travel time at time $p_{ik}$ divided by the free flow travel time, which is commonly used in the literature [24]. With this index, a routing schema which minimizes the TTI does not necessarily minimize the travel time. This is due to the fact that an optimal routing plan may select a longer route in order to avoid the congested roadways, roadways with lower speed limits, or a combination of both. In fact, Objective (2) is formulated to provide a more uniform driving pattern with less cars on the roadways even if it requires longer travels in term of distance and time.

Constraint sets (3-7) are graph construction constraints, and are defined as follows: Each customer node must be visited exactly once (3), and maximum of $K$ vehicles can be used for the routing plan (4). This constraints allow the model to utilize the fleet partially, and therefore keep some vehicles inactive in the depot. Dispatched vehicles must return to the depot after serving (5). When a vehicle enters a customer's node, it will leave after serving (6). Nodes $v_0$ and $v_{n+1}$ are defined for fleet dispatching and return from/to depot (7). Finally, vehicles have certain capacities that cannot be exceeded (8). Constraint sets (9-11) are timing constraints: Service should be performed within the pre-defined time windows at each customer node (9). Constraints (10) tracks the arrival/departure times from/to each node, and also guarantees that routing schedule does not exceed the latest planning time. Constraints (11) allow vehicles to stay at nodes in order to avoid traffic congestion. Similar to the parameters in objective function, travel time $t_{ijk}(p_{ik})$ is a function of time of day and the average speed at that certain time. While solving the VRP problems, sub tours are one of the main difficulties encountered. However, unlike traditional VRP formulations, we do not provide subtour elimination constraints. In the safe VRP formulation presented herein, sub tours are eliminated through the collaboration of constraints (10) and (11).

Please also note that we consider 24 hours of the day in the model. Let $T$ denote each hour of day $T=\{T_1,...,T_m\}$, where m=24. Each $T_l$ has an associated speed $h_l$, where $l$ is between 1 and m. The earliest and latest moment in each time interval in defined as $t^l_{min}$ and $t^l_{max}$. Depending on the length of an arc, hour of day, and also the speed at that hour, a vehicle may face different levels of traffic flow, and consequently speed variations. In other words, a vehicle travels on one arc in one or more time intervals. We demonstrate a set of time periods from node $i$ to $j$ for vehicle $k$ as $t_{ijk}(p_{ik}) = \{ t^l_{ijk}, t^{l+1}_{ijk}, ... \}$, and travel speeds $S_{ijk}(p_{ik}) = \{ S^l_{ijk}, S^{l+1}_{ijk}, ... \}$. This is used in the proposed model, and our algorithm is capable of incorporating this concept.

We compare the results of the proposed approach to the classical network optimization objectives of minimum distance minimum travel time and, defined, respectively, in functions (15) and (16):

$$minimize \ \sum_{k=1}^{K} \sum_{(i,j)\in V} d_{ijk}\, x_{ijk} \tag{15}$$

$$minimize \ \sum_{k=1}^{K} \sum_{(i,j)\in V} [s_i + t_{ijk}(p_{ik})]\, x_{ijk} \tag{16}$$

*2.2. Phase 2: Speed and departure time scheduling*

In the route scheduling literature, including ship routing and scheduling, air cargo or other transit scheduling models, it is assumed that are routes are predetermined, and optimization approaches search for the optimal speed and scheduling [25-27]. The Phase 2 of the proposed approach, based on the graph and routes constructed in Phase 1, solves a second optimization model which considers each route as an independent transit path (fixed route with fixed node sequences) [25] and determines the service timing to avoid congestion by rescheduling the departure times of each vehicle from each node and finds the optimal speed on each arc. Fig. 1c illustrates the scheduling decisions made in Phase 2. Therefore, Phase 2 is inspired by this type of approaches.

We employ the concept of shortest path for speed and departure time optimization [25]. Each node in Fig. 1c is expanded to several nodes corresponding to different scenarios (arrival time to the successor node). In other words, the arrival time (within the time window) at each node are discretized, and on the generated directed acyclic graph for each route, our aim is to find the minimum cost. Arrival time at a node depends on the departure time from the previous node, and the speed on the arc connecting each pair of nodes. Hence, the shortest path (minimum cost) should have the optimal departure time and speed values. Each discretized arrival time scenario is represented as $N_{is}$ where i is the node number and s is the number of scenario $\left(N_{i,s+1} \geq N_{i,s}\right)$. Every arc $((i,s),(i+1,p))$ in the shortest path graph connects a discretized arrival time for node s to a discretized time for node s+1 with the cost defined as $c_{(i,s),(i+1,p)}$,. For each node in route k, we define m as the number of discretization. The variable $x_{(i,s),(i+1,p)}$ takes a value of 1 if arc $((i,s),(i+1,p))$ is used, and 0 otherwise. The mathematical formulation of the model proposed in Phase 2 is as follows:

Objectives:

$$Minimize \sum_{i=1,\dots,n-1} \sum_{s=1,\dots,m} \sum_{p=1,\dots,m} C_{(i,s),(i+1,p)}\, x_{(i,s),(i+1,p)} \tag{17}$$

Subject to:

$$\sum_{s\in N_i} \sum_{p=1,\dots,m} x_{(is),(i+1,p)} = 1 \; ; \quad i = (1,\dots,n-1) \tag{18}$$

$$\sum_{s=1,\dots,m} x_{(i-1,s),(i,p)} = \sum_{s=1,\dots,m} x_{(i,s),(i+1,p)} \quad ; \quad i = (1,\dots,n) \quad , \quad p = (1,\dots,m) \tag{19}$$

$$x_{(is),(i+1,p)} \in \{0,1\} \quad ; \quad i = (1,\dots,n-1) \quad , \quad s = (1,\dots,m) \quad , \quad p = (1,\dots,m\,) \tag{20}$$

Phase 2 objective function (17) seeks the shortest path that provides the minimal cost. In our model, we try to minimize the crash risk and congestion. Constrains set (18) guarantees all the transit nodes (nodes in route) are served, and the flow in acyclic graph of shortest path formulation is evaluated through constraint sets (19). The Phase 2 optimization model is run for each route obtained from Phase 1. In the following sections, we will discuss that our proposed algorithm solves Phase 1 and then Phase 2 iteratively and consecutively in each iteration to find an optimal routing and scheduling plan.

*2.3. Queuing based dynamic speed prediction*

In sections 2.1 and 2.2, we discuss that variable speed values are crucial due to the dynamic logic of our models. Numerous agencies and companies collect such real-time speed (Inrix, Total Traffic, etc.). In addition, navigation devices utilize this data to navigate users. Therefore, the proposed models can be fed to these navigation and routing devices to obtain and plan for safer routing. However, real-time speed (or travel time) data is not always available. Considering an average speed value on arcs also cannot represent the real congestion or traffic conditions. In the literature, many researchers considered a step function to estimate speed and travel times. However, studies have proven that it cannot efficiently represent real-world patterns [8, 9, 28]. Therefore, in this study, a queueing model is used to calculate and estimate speeds, and consequently the travel times on arcs. For queue modelling, traffic flows on every path is needed. We calculate speed on every path at each hour of the day by using the concepts of the well-known traffic flow theory [29, 30]. This equation relates the traffic flow ($Fijy$), traffic density ($K$ijy) and speed ($Sijy$) between nodes $i$ and $j$ in hour $y$.

$$Fijy = Kijy \times Sijy \tag{21}$$

Queueing parameters are listed in Table 2. We split a roadway segment into multiple segments with a length of $1/K_{jam}$, where $K_{jam}$ is the maximum traffic density (this is the length that a vehicle occupies in a path) [31]. Each segment is assumed as a service station, where cars arrive at the rate of $\lambda$, and get a service with rate of $\mu$. We formulated an M/M/1 model, but the performance of speed prediction was not satisfactory. However, since the distribution of time between arrivals has been proven to follow a Gamma family distribution [9, 32], we kept the arrival rate as Poisson, and changed the service time to general distribution (M/G/1). Hence, the effective speed can be obtained by dividing the segment length ($1/K_{jam}$) over the total time spent in the system ($W$). Also nominal speed can be considered as the posted speed limit, the value 5 added to the posted speed limit, 85$^{th}$ percentile of the speed observations, free flow speed or any other speed level depending on the characteristics of the site. One may need to pay close attention to the selection of the speed as it can influence the drivers' choice of speed and consequently the system. For further information on the choice of speed please refer to [33]. In this study we considered the value equal to the posted speed limit. From this one can calculate the unit less relative speed by dividing the actual (effective) speed ($Sijy$) over the Nominal Speed.

Table 2 M/G/1 Queueing Parameters

| Parameter | Definition | Unit |
|---|---|---|
| $K$ | Traffic density | Vehicle per mile |
| $K_{jam}$ | Maximum traffic density | Vehicle per mile |
| $S$ | Effective speed | Mile per hour |
| $S_N$ | Nominal speed | Mile per hour |
| $R$ | Relative Speed | --- |
| $F$ | Traffic flow | Vehicle per hour |
| $W$ | Total time spent in the system | Hour |
| $\lambda$ | Arrival rate | Vehicle per hour |
| $\mu$ | Service time | Vehicle per hour |
| $\rho$ | Traffic intensity | $(\lambda/\mu = E/K_{jam})$ |

Vandale et al. (2010) formulated the waiting time and relative speed for general and several special cases. Here, we explain the formulation for M/G/1 in detail. The service time in this model is generally distributed with a service time of $1/\mu$, and a standard deviation of $\sigma$. Hence, expected service rate is $\mu$, and is calculated as $\mu = S_N \times K_{jam}$.

**Lemma 1**. Total waiting time for the M/G/1 queuing system is:

$$W = \frac{1}{S_N C} + \frac{\rho^2 + S_N^2 K^2 \sigma^2}{2 S_N K (1-\rho)} \tag{22}$$

**Proof 1**. For the general case, $W = \left(1/K_{jam}\right)/S$ and $W = 1/\mu - \lambda = 1/\left[S_N(K_{jam} - K)\right]$. Combining Little's theorem and the Pollaczek-Khintchine formula for the average number of cars in the system, and substituting for $\lambda$ and $\mu$, total waiting time in the system is calculated [34].

Using Equation (22), speed and relative speed are calculated as:

$$S = \frac{2S_N\left(K_{jam} - K\right)}{2K_{jam} + K(\beta^2 - 1)} \quad , \qquad R = \frac{S}{S_N} = \frac{2(\rho - 1)}{2 + \rho(\beta^2 - 1)} \tag{23}$$

where $\beta$ delineates the coefficient of variation of service time, and calculated as follows: $\beta = \sigma S_N C$. Finally, flow-density-speed function can be obtained by substituting (21) in (23):

$$f(S, F) = 2K_{jam}S^2 + \left[F(\beta^2 - 1) - 2K_{jam} \times S_N\right]S + 2F \times S_N = 0 \tag{24}$$

The shape of the flow-density-speed curve changes by assigning different values to $\beta$. For a moderate variability in service time, $\beta = 1$ seems to be reasonable. In the case study section, we will assume $\beta$=1; however, the methodology can be applied to cases where $\beta$ has a different value. After these curves are created, we select the threshold speed from the curve and calculate dynamic travel times. Please refer to the Case Study section for the selection of the speeds from the curves. Travel time values are stored in a hyper-matrix, and are fed to the models according to the time of day. Next section discusses how the proposed algorithm solves the models using these data.

## 3. Solution Approach

VRP is a NP-Hard problem and by a linear growth in the size of problem, the calculation time and complexity of model increases exponentially. Therefore, heuristic and meta-heuristic algorithms are developed to tackle the complex problems. In our VRP formulation the first component of the objective function is nonlinear. Our first model is also not deterministic, and the parameters change according to the time of day. As such, it is not feasible to solve them through optimality using exact algorithms. Hence, in this study, in order to achieve a fair tradeoff between the computation time and solution accuracy, we propose to solve the aforementioned problem using a hybrid algorithm, which combines a novel heuristic algorithm and a sophisticated meta-heuristic technique. A schematic approach of the solution approach is depicted in Fig. 2. For the example given in Fig. 2, our proposed initial solution generator algorithm divides the plane into five slices, with the center of D and the angle of $2\pi/5$. Each slice is further divided into two sub-slices. Therefore, there will be ten slices of equal angle. Our algorithm starts from the closest node to D, and adds the nodes in the slice $2i-1$ to vehicle $i$ with the order of closest node to D to the furthest one, in terms of direct distance. For even slices, the procedure is the same, but the order of nodes is from the furthest to the closest node to depot. Finally, the last node of region $2i-1$ is joined to the first node of $2i$, a 2-opt neighborhood search is conducted on each of 5 slices, and the initial solution is obtained as in Fig. 2-a. The pseudo-code of this algorithm is provided in Fig. 3 for further detail.

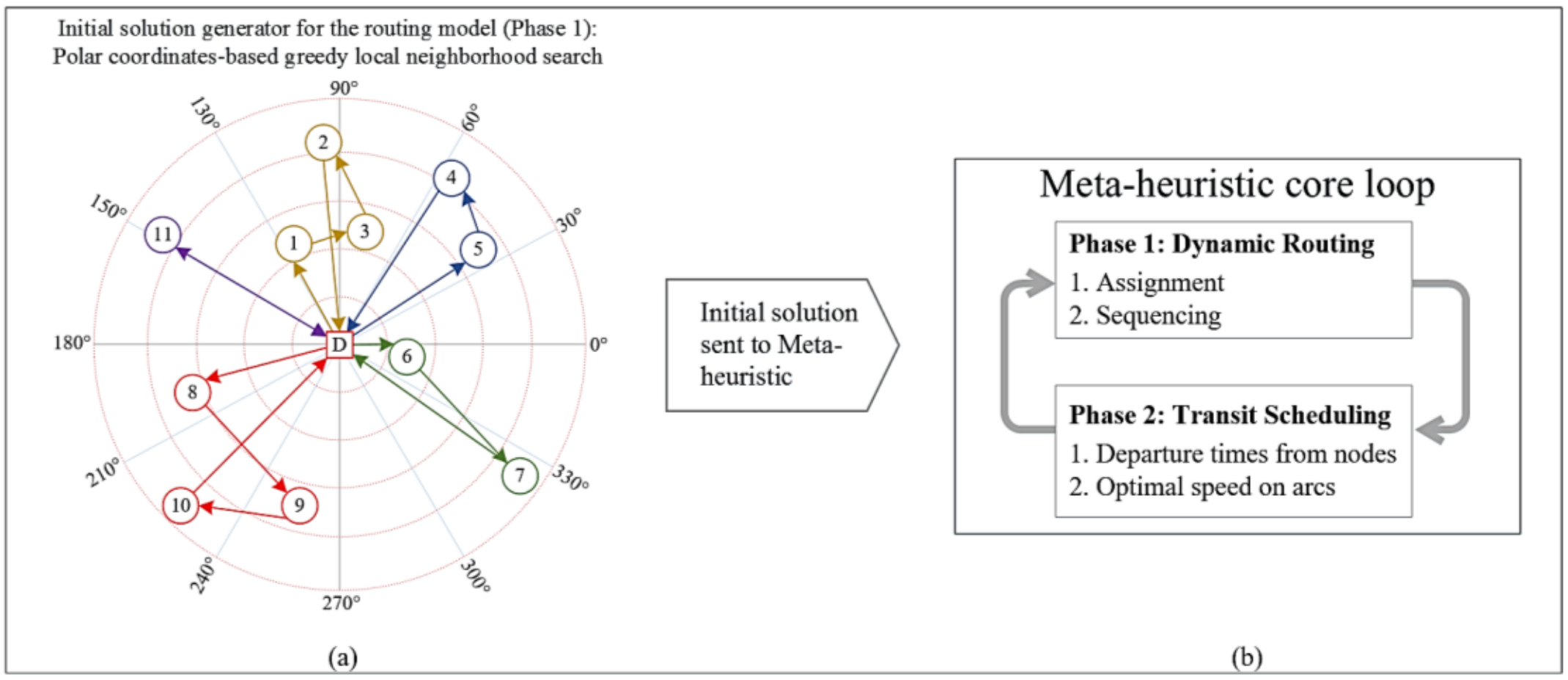

Fig. 2. A schematic representation of solution approach (a) Polar coordinates-based heuristic to generate the initial solution (b) Meta-heuristic to solve Phase 1 and Phase 2 problems

An enhanced simulated annealing (SA) algorithm is the meta-heuristic approach employed in this research. In the SA, developed by Kirkpatrick [35], the acceptance probability of an inferior solution is defined as $exp\ (-\Delta f/t)$, where $\Delta f$ denotes the solution gap between the current and neighbour objective function values, and $t$ is the temperature variable in SA. Equations (25) and (26) shows these parameters (This is known as Boltzman criterion) [35].

$$Acceptance\ rate\ of\ s'\ (against\ s) = \begin{cases} e^{-\frac{\Delta f}{t_k}} & \Delta f \geq 0 \\ 1 & else \end{cases} \tag{25}$$

$$\Delta f = f(s') - f(s) \tag{26}$$

We also define the cooling factor in Equation (27). This formulation indicates that if we start from $T_0$ with this α , we get to $T_0$ at the end or iterations on outer loop of SA. For local neighborhood search procedure, we introduce 6 categories of heuristics in SA: (1) insertion, (2) swap, (3) 2-OPT, (4) 3-OPT, (5) Reversion, (6) Split [36]. The first two have two sub heuristics depending on the number of nodes selected to be inserted or swapped.

$$\alpha = \left(\frac{T_f}{T_0}\right)^{(1/max.\ iteration)} \tag{27}$$

This algorithm starts from a random initial solution, and the final solution quality depends on the initial solution. Therefore, in order to obtain a good initial feasible solution to the problem, we use the heuristic algorithm developed by the authors [13]. This approach uses a greedy local neighborhood search algorithm that works with polar coordinates of the nodes, where the depot is the initial pole, and expands the search range along the radius and azimuth, successively (Fig. 2a, Fig. 3). The initial solution is then fed to the meta-heuristic algorithm, which develops itself to improve the solution quality at each iteration leading to suboptimal solutions.

```
Algorithm: Polar coordinates-based local search
Input: k , ε , n, max(d) ;
Output: Perm.(nodes)on k^th vehicle*

do: initialization
  ∟ Input network of nodes and depot

divide the plane into k equal slices
    for i  in {k}
        if i is 1 do
                Lower limit angle (i) = ε
                Upper limit angle (i) = (2π/k) + ε
        else do
                Lower limit angle (i) = ε + (2π/k).(i − 1)
                Upper limit angle (i) = ε + (2π/k) i
        end
        do Split each slice with bisection line
    end
    temp_odd={empty}
    temp_even={empty}
    for i in {1: 2: 2k − 1}
        for j in {1: max(d)/n}
        Increase local search region (0-j)
(1) a.  Azimuth
    b.  Radius
        do assign nodes in search region i to temp_odd [(i + 1)/2].
    end
    for i in {2: 2: 2k − 2}
        for j in {1: max(d)/n}
        Increase local search region (0-j)
(2) c.  Azimuth
    d.  Radius
        do assign nodes in search region i to temp_even [i/2].
    end
    Concatenate permutations [(1), (2)]

k: Number of active vehicles
ε: Initial Angle
n: Seach expansion step function (n cocenteric circles)
max(d) : Maximum direct distance of customer nodes from depot
```

Fig. 3. Pseudo code for heuristic initial solution generator: Polar coordinates-based greedy local neighborhood search

## 4. Model Evaluation on Benchmark Problems

We generate several small, medium and large scale instances employing the Solomon logic as well as its extant benchmark instances for the proposed VRP with time windows model evaluation. This benchmark set is one of the most commonly used benchmark instances in VRP in the context of time windows. For further research on Solomon's methodology, please refer to [37]. In detail, demand, service time, time windows, and the position of nodes and depot is modelled according to the Solomon's benchmark networks. Crash probabilities, travel time indices and speeds on are generated according to a step function of three levels and five intervals. Note that crash probability and travel time index values increase during rush hours whereas the speed drops. Next, we inject a moderate uniformly distributed noise to differentiate these parameters on different arcs. In other words, this noise is introduced to the data to reflect the real-world stochasticity more accurately than a deterministic step function. This noise does not interfere with the time-dependent nature of the model and is only introduced to randomize the values from one time interval to another. The amount and the direction of the noise is simulated based on our observation on the real world case. Fig. 4 shows an example of this approach for one arc. Dashed line in Fig. 4 is the initial step function for the parameters whereas the solid line is the noisy distributions. Noise is introduced to all 24 hours; however, the proportion of the noise is randomly chosen. Fig. 4a shows high noise proportion for crash probability whereas it is low for TTI, as seen in Fig. 4b. Since TTI and speed are correlated, the proportion of noise introduced for these two parameters on each arc is similar.

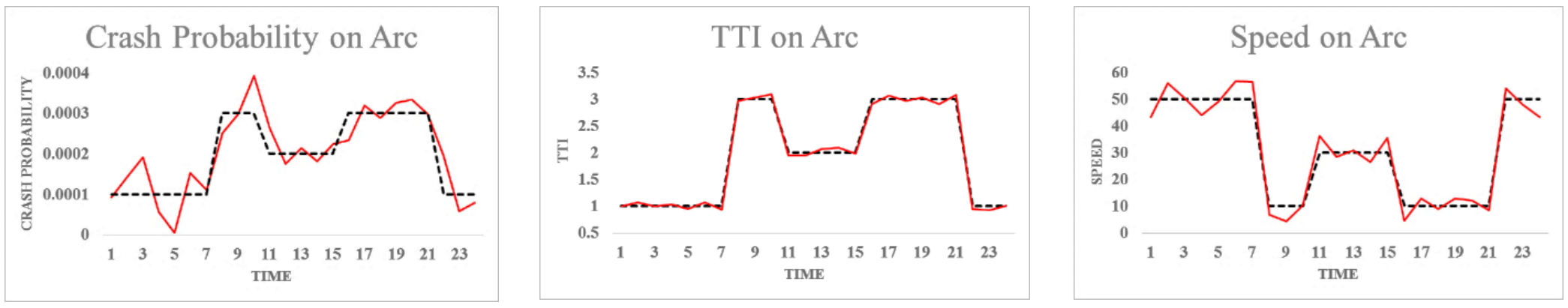

(a) (b) (c)

Fig. 4. Variations in Crash Probability, TTI and Speed Introduced as Noisy Step Functions. Dashed Line Is the Step Function before Introducing the Noise, and Solid Line is the Noisy Data (Input to the Test Problems).

For parameter tuning, we employed a Latin hypercube sampling design. A total of 30 input parameter configuration is created for each scenario, and they are tested 5 times in the algorithm where the minimum value is returned as the objective function value. Inferences are performed with a Gaussian process regression, namely Kriging, using the JMP statistical analysis software [38, 39]. The most accurate parameter tuning is selected among the all kriging outputs, and the best performance out of 30 configuration. All experiments are conducted on an Intel core i7-5500U 2.4 GHz processor, 8 GB Ram, Win 10 Pro platform. Table 3 shows the performance of our algorithm. Although the results for the main objective function is also presented in Table 3 as a reference, our main objective is to compare the minimum travelled distance obtained from our model to the best lower bounds obtained by Solomon (1987).

Table 3. Model Validation and Algorithm Performance Assessment. R Refers To Benchmark Series R (Randomly And Uniformly Distributed Nodes). First 4 Rows Are Generated Instances, Rows 5 to 10 Are Instances Taken from Solomon's Test Problem Dataset. Abbreviations: Shortest Path (SP), Number of Vehicles (#V), Travelled Distance (TD), Computational Time in seconds (T), Objective Function Value (OFV), Objective Function Gap in percent (OG), Computational Time Gap in percent (TG).

| | Exact/LB | | Basic SA | | | Augmented SA | | | | Comparison | |
|---|---|---|---|---|---|---|---|---|---|---|---|
| Instance | SP | #V | OFV | #V | T | OFV | TD | #V | T | OG | TG |
| R10 | --- | --- | 1.62 | 2 | 272 | 1.62 | 244.4 | 2 | 65 | 0.0 | -318 |
| R25 | --- | --- | 1.59 | 2 | 322 | 1.59 | 311.0 | 2 | 121 | 0.0 | -166 |
| R50 | --- | --- | 1.66 | 4 | 419 | 1.58 | 973.5 | 3 | 207 | -5.1 | -102 |
| R80 | --- | --- | 1.66 | 7 | 601 | 1.60 | 1366.5 | 7 | 359 | -3.7 | -67 |
| R101 | 1645.7 | 19 | 1.71 | 20 | 982 | 1.59 | 2462.6 | 20 | 542 | -7.5 | -81 |
| R105 | 1377.1 | 14 | 1.73 | 20 | 949 | 1.66 | 2004.2 | 20 | 530 | -4.2 | -79 |
| R109 | 1194.7 | 11 | 1.72 | 20 | 927 | 1.63 | 2044.7 | 20 | 561 | -5.5 | -65 |
| R201 | 1252.3 | 4 | 1.72 | 19 | 846 | 1.67 | 2515.0 | 15 | 474 | -3.0 | -78 |
| R205 | 994.4 | 3 | 1.74 | 18 | 872 | 1.70 | 1243.5 | 14 | 465 | -2.4 | -88 |
| R209 | 909.1 | 3 | 1.33 | 19 | 829 | 1.10 | 1196.4 | 12 | 438 | -20.9 | -89 |

The first four rows are the test problems generated using the Solomon's approach, for 10, 25, 50 and 80 nodes, respectively. Fleet sizes for these instances are 2, 3, 5, 12 vehicles, respectively. The time windows are considered as hard time windows on upper bound and soft on lower bound. That is, early arrival is allowed; however, service cannot be started until the lower bound. On the other hand, late arrival at nodes is not allowed. Due to the NP-complete nature of the models, solving the instances to the optimality is not feasible. Therefore, the cells for these instances are left blank. These instances Benchmark test problems R100 series refer to those with hard time windows, and R 200 series are those with semi-relaxed time windows. Dispatching time is set to 7:00 AM. In terms of the computational time, we observe that the proposed algorithm is capable of solving the test problems is a relatively short period of time. Note that the travelled distance returned by our approach is considerable higher than the lower bound. This observation is not counter-intuitive since the objective function is to obtain the safest path rather than the shortest path in this paper. In the following sections, we will discuss the trade-off between the safe routes based on the travelled distance. The computation time results also indicate that the model is capable of finding high quality results within an acceptable time period even for large and complex network problems. We compared the results from our proposed algorithm to those obtained from the basic SA algorithm as described by Kirkpatrick [35]. It has been shown in the literature that SA performs well for similar models [4, 40]. Our proposed approach was able to increase the solution quality by 6 percent (on average), while the computation time has drastically decreased between 70 to 320 percent.

## 5. Case Study

Safe vehicle routing approach proposed in this paper is applicable in many practical situations, including routing for dial-a-ride services for aging population, transportation of sensitive goods or dangerous substances, and ambulance routing. In this section we illustrate the application of the proposed models on a real-world case study. We study the ambulance routing problem, and consider the following four hospitals in the City of Miami to be served with one vehicle:

0. University of Miami Hospital (Central Depot)
1. The Miami Medical Center
2. North Shore Medical Center
3. Mount Sinaj Medical Center

Fig. 5 shows the locations of the hospitals (nodes), and gives the graph definition of the model along with the distance matrix. We choose the material to be transported as blood units in this case study. These blood units are to be dispatched from the main depot to other three demand nodes. Our objective is to select an optimal route and departure schedule that minimize both the risk of congestion and the risk of crash. Each demand node has time windows and certain demand, and vehicles stop at each node to serve them. Table 4. Parameter Values for Nodes in Case Study lists the parameter values used in the problem. In Fig. 5, the distances are city block shortest distances. For the sake of simplicity, we transform our real-world network into a graph (Fig. 5-right). Arcs between the depot and demand points (1, 2, 3) are major roadways including primary arterials and interstate roadways. Shortest distance arcs (4, 5, 6) are shown on Fig. 5-left between each pair of demand points, which are mostly collectors and local roadway segments.

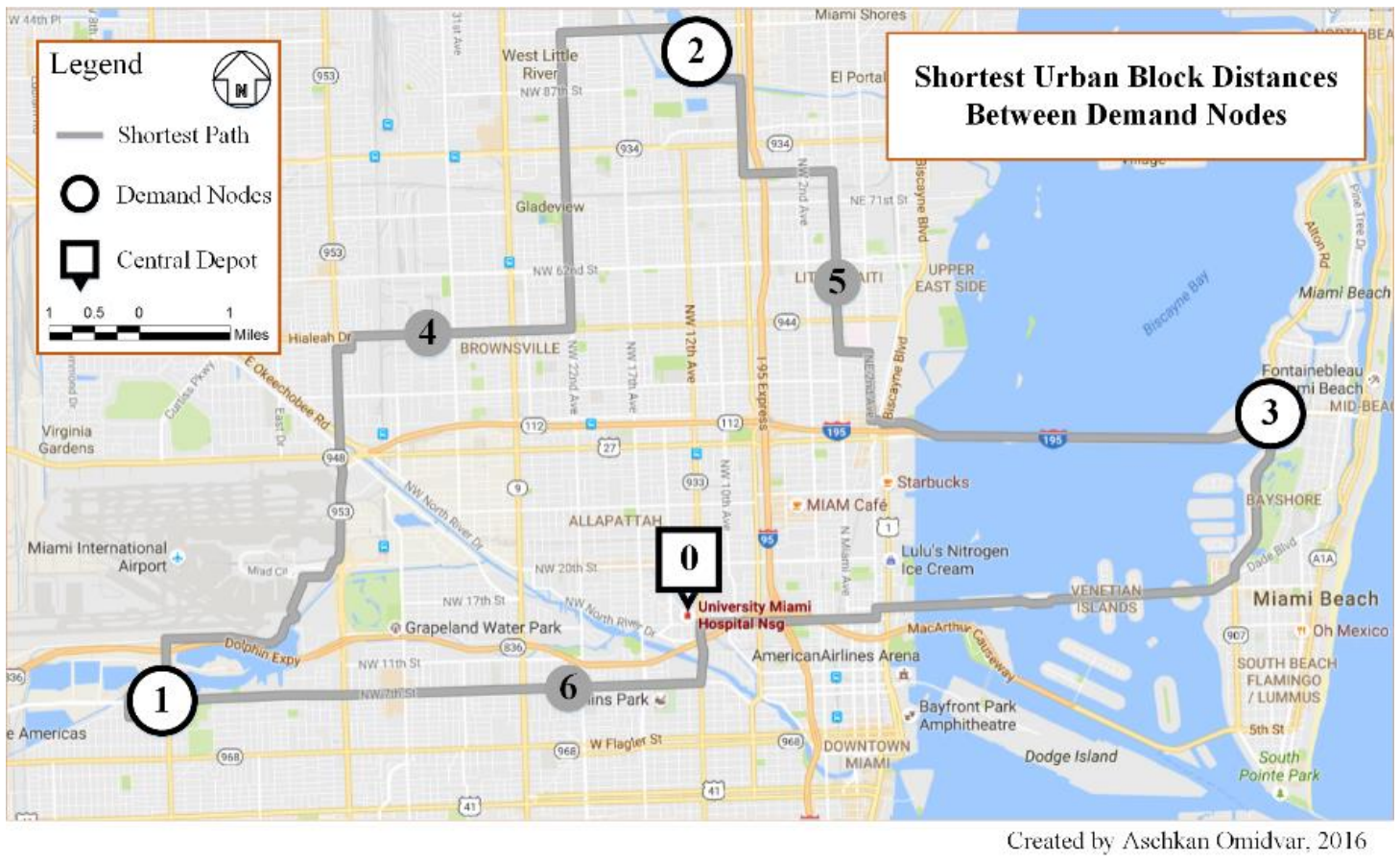

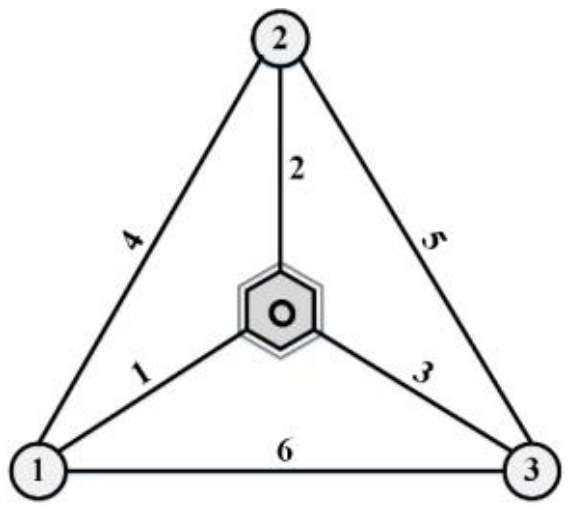

O-D distance matrix

| | 0 | 1 | 2 | 3 |
|---|---|---|---|---|
| **0** | **0** | **6.898** | **5.500** | **6.402** |
| **1** | **6.898** | **0** | **10.202** | **12.003** |
| **2** | **5.500** | **10.202** | **0** | **8.800** |
| **3** | **6.402** | **12.003** | **8.800** | **0** |

Fig. 5. (Left) Real-World Network for Safe Routing Study – Miami. (Right) Network Transformed to Graph and Urban Block Distance Matrix

In the transformation of city map (Fig. 5-left) to a graph (Fig. 5-right), the Hamiltonian graph condition is violated. In other words, although each node can be visited once in a VRP graph, we assume that a vehicle may use a roadway more than once to get access to an unserved node in our formulation. For instance, suppose that the vehicle served the node 3, and is now at node 1. The vehicle's final destination is the node 2, and the only possible option is arc 4 given the Hamiltonian graph representation. However, in the real world, a vehicle may also use the arc 1, and therefore can get back to the node 2 through arcs 1 and 2 without entering the main depot. We fix this issue with the following Proposition 1. This idea is similar to the concept of VRP with satellite facilities introduced by Bard et al. [41], and alternative fuel stations by Erdogan and Miller-Hooks [42].

**Proposition 1**: An arc can be used more than once in a VRP network.

**Proof**: For this purpose, multiple visit should be allowed to a node. Let's augment the current graph to a new one as $G' = (V', E')$ with new dummy vertices for the main depot, $W = \{v_{n+2}\}$, where the new vertex set is defined as $V' = V \cup W$. The number of dummy vertices associated with $v_{n+2}$, $m$, is set to the number of visits. $m$ should be small enough to reduce the network size, but large enough to enable the full utilization of the roadway network.

Table 4. Parameter Values for Nodes in Case Study

| | Parameter | description | Value |
|---|---|---|---|
| 1 | $s_i$ | Service time at node $i$ | $\{0.1, 0.1, 0.1\}$ |
| 2 | $q_i$ | Demand at node $i$ | $\{100, 120, 80\}$ |
| 3 | $Q_i$ | Capacity of vehicle $i$ | 300 |
| 4 | $[e_i, l_i]$ | Time windows on node $i$ | $[e_i = 0, l_i = 1.3]$ |
| 5 | $L$ | Latest Departure Time | # of scenario-1 |
| 6 | $m$ | Dummy vertices associated with depot | 2 |

In order to formulate the objective functions of Eq. (1), we use the hourly flow data and average hourly speed data. The data is collected on telemetric traffic monitoring stations (TTMS). For all arcs that have a TTMS station, we obtain three years of hourly flow data between the years 2010 and 2012. We use all the data points to generate the traffic flow, density and speed relationship diagrams. For the arcs with no flow data at hand, we perform simulation based on roadways in vicinity, where flow data is available. Researchers are encouraged to use predictive methods and crash analytics to estimate the parameters for the objective function [43].

Fig. 6a and Fig. 6d show the average hourly flow for Arc 3 in the eastbound and westbound directions for the selected three years. Following the equations (21)-(24), we create the flow, density and speed relationship diagrams as depicted in Fig. 6b, Fig. 6c, Fig. 6e and Fig. 6f. Nominal speed ($S_N$) is set equal to the free flow speed on each arc/direction. Assuming $\beta = 1$ for the graph and by taking the derivative of Eq. (24) maximizing for $s$, we obtain the maximum traffic flow for M/G/1 as follows, where the problem reduces to the well-known case of linear relationship between the speed and density:

$$F_{max} = \frac{S_N \times K_{jam}}{4} \tag{28}$$

For each arc and direction, the highest flow value (assuming normal environmental and traffic conditions) is set to $F_{max}$. From Eq (28), we calculate the jam density, and create flow, speed and density relationship diagrams shown in Fig. 6. As shown in Fig. 6a and Fig. 6b, for each flow value, we have two speed values that correspond to congested and uncongested regimes. Depending on the classification of the roadway, we set our thresholds between 0.854 and 0.901 quantiles of flow (see [44]) for default values) to determine the level of congestion. Congested speed values are those corresponding to the flow values above the threshold, and uncongested speed values are those corresponding to the flow values less than the selected threshold. Fig. 7 illustrates the performance of our approach and M/G/1 queuing model in simulating the speeds at every hour. We observe that estimated speeds have a good concordance with the real-life data. We repeat the procedure for all arcs and directions, and feed this data into the optimization models. In the models, speed data is used to calculate the travel times and TTI.

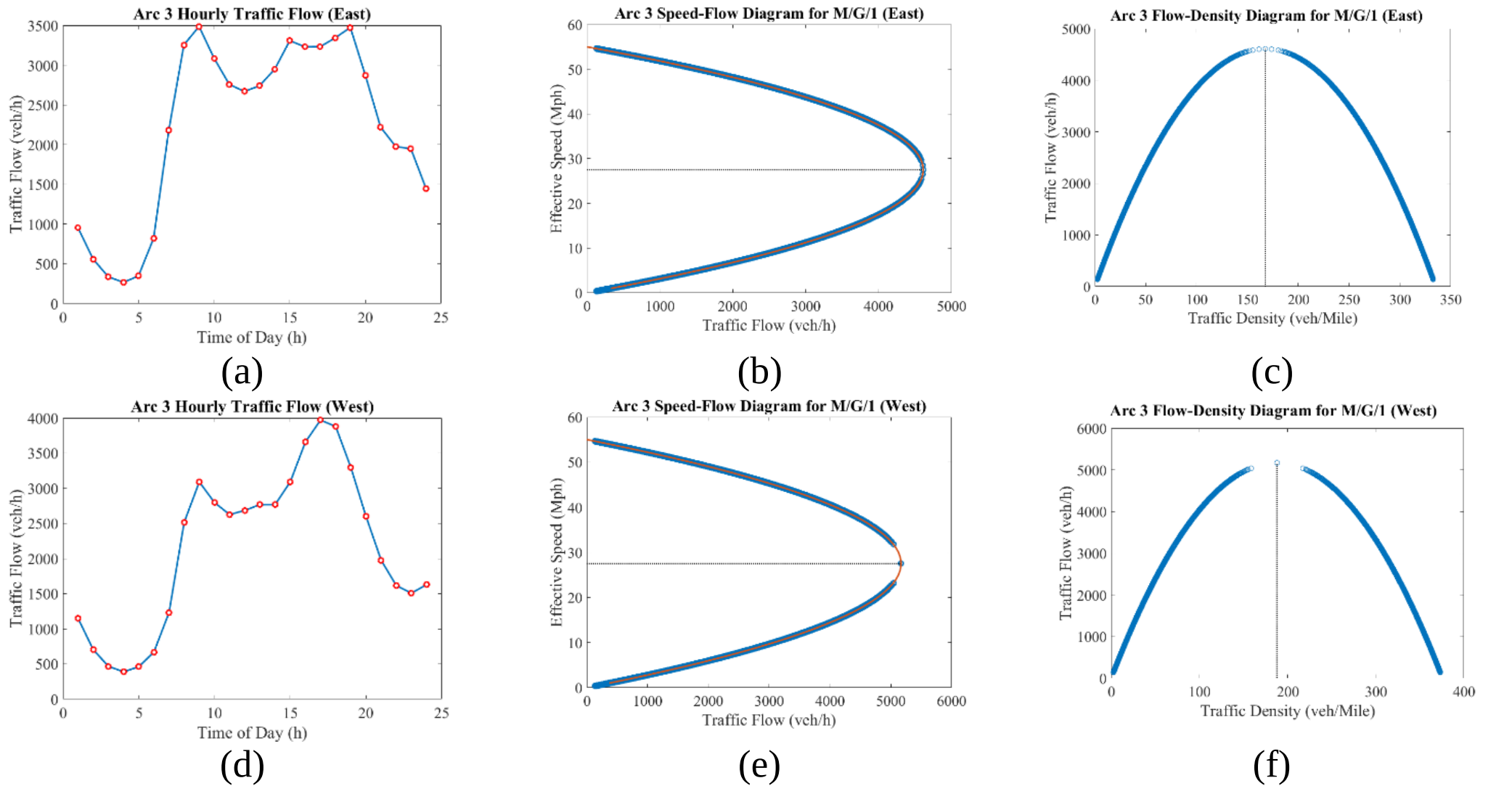

Fig. 6. An Example of Flow Analysis in The Case Study on Arc 3 West and East Directions

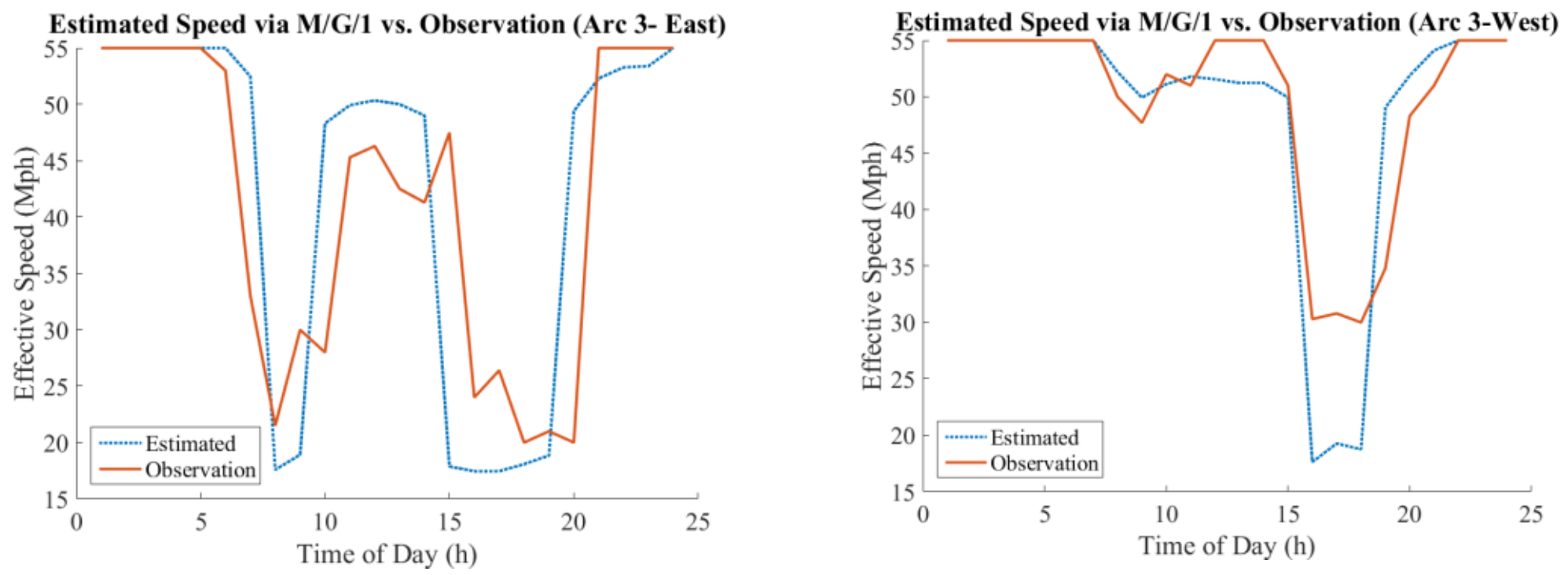

Fig. 7. An Instance of Speed Simulation on Arcs Using M/G/1 Queuing Approach on Arc 3 (East and West)

To determine the crash probability on each arc, past crash history are obtained from the Traffic Safety Office of the Florida Department of Transportation (FDOT) between years 2010 and 2012. Through an extensive GIS analysis, we classify crashes on each arc according to the direction, and put the crash counts in hourly intervals. For example, Fig. 8 shows the hourly crash characteristics on the arc 3 for both directions. For some arcs, depending on the roadway class, crash count increases during rush hours, and for some others this increase is negligible. These differences are reflected in the routing plans obtained by the optimization models. Since arc lengths are not equal, we normalize the hourly crash counts by arc length to get the number of crashes per mile for each arc. The normalized crash counts are divided by the total number of hours in three years to calculate the crash probabilities. Finally, we scale the probabilities in order to have the same order of magnitude for both the normalized TTI and the crash probabilities.

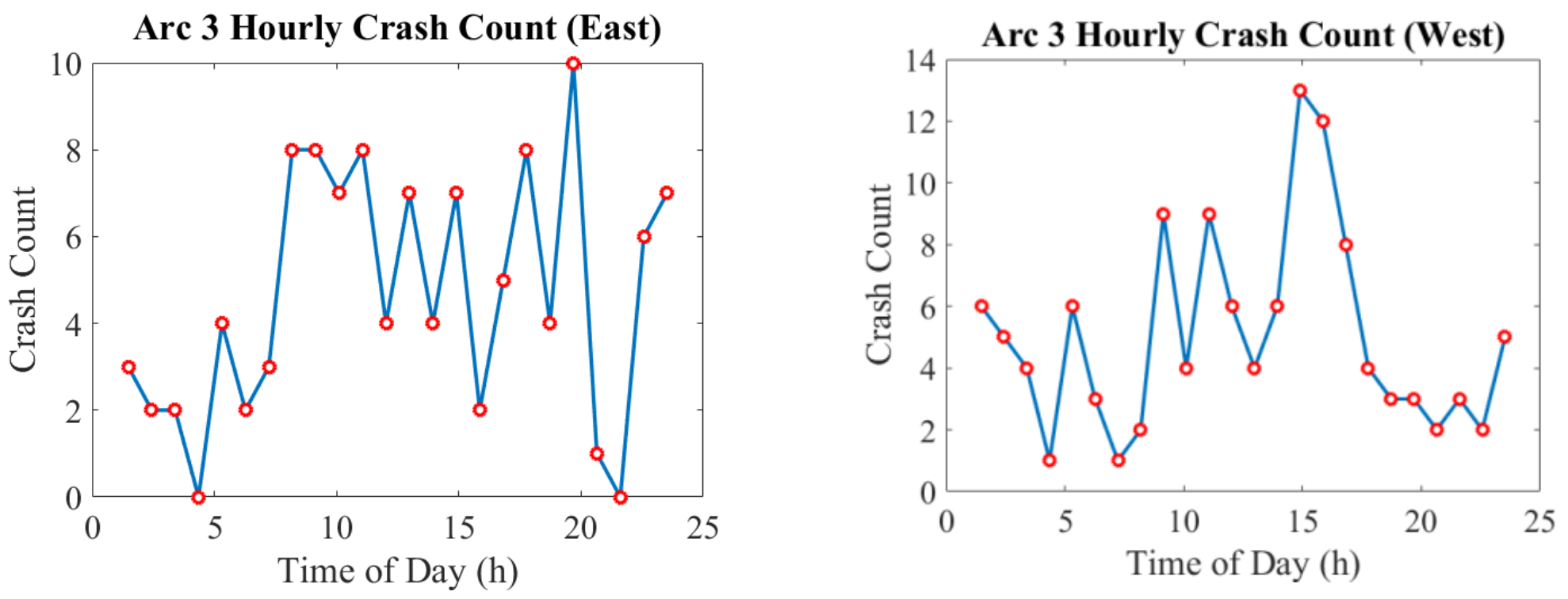

Fig. 8. Hourly Traffic Crash Patterns on Arc 3 (East and West)

A total of 24 scenarios are considered here. A total of 24 scenarios are considered here. Starting from 12 AM, a new scenario is created every hour thereafter for the next 24 hours. We conduct exhaustive enumeration on all scenarios for the case study in order to check the performance of the proposed augmented SA approach. This approach was able to solve both models to global optimality in less than two seconds. The following parameter values are selected for the proposed parameter tuning:

- Maximum number of iterations: 10
- Initial temperature: 10
- Final temperature: 0.01
- Maximum number of iteration per temperature: 5
- Number of population admitted to next iteration: 4
- Cooling factor: Boltzman index

The proposed optimization approach minimizes the weighted sum of the functions given in Eq. (1) and Eq. (2) to find the safest route with least congestion. We compare this solution to the safest route and the least congested route solutions obtained, respectively, by minimizing Eq. (1) and Eq. (2) separately. In addition, we compare the optimal solution to the following: (a) the fastest route (minimum travel time) obtained by considering (16) as the objective function, and the shortest route (minimum travel distance) by minimizing (15). Table 5 gives the minimum travel time routes. It can be seen that the slowest route occurs at 5pm interval with 61.20 min. The minimum travel distance route does not depend on the time of the day. Minimum distance solutions found as 3-5-4-1 or 1-4-5-3 with the total travelled distance of 32.30 miles.

Table 6 summarizes the routes obtained from the proposed method for each time interval. In Table 6, optimal routes are enumerated according to arc permutation (not nodes). Our modelling approach is node-based optimization; however, we use the arc numbers for better readability and ease of tracking the routes. Minimum distance is achieved through the following routing plan: 3-5-4-1 or 1-4-5-3 with the total travelled distance of 32.3009 miles. The results with objective function (16), where only the travel time is minimized, indicates that when the speed is close to free flow speed, routing is planned on interstate highways (arcs 1, 2 and 3) and in other hours, routes change (Table 5). During rush hours specifically, routing is performed through a combination of major and local/connector roadways. Routing with the minimum travel time is the same as routing with minimum distance during congested hours. Finally, crash risk cost gap indicates that the minimum travel time routing plan is considerably different from the safest path routing plan.

Columns of Table 6 present the output routing plans of three models classified as: "Weighted Sum of Objectives" is the weighted sum of Eq. (1) and Eq. (2), "Crash Risk Minimization" associated with Eq. (1) only, and "Congestion Avoidance" associated with Eq. (2).

Table 5. Minimum travel time routes

| Start Time | Route | OFV (min.) | Crash Risk Cost Gap | Traversed Distance Gap (Compared to min distance) |
|---|---|---|---|---|
| 12:00:00 AM - 6:00:00 AM | 1-1-2-2-3-3 | 41.02 | > 0 | 5.299 |
| 7:00:00 AM | 1-6-3-2-2 | 46.76 | 0.149 | 4.002 |
| 8:00:00 AM | 1-6-3-2-2 | 47.29 | 0.154 | 4.002 |
| 9:00:00 AM - 1:00:00 PM | 1-1-2-2-3-3 | 44.22 - 44.95 | > 0 | 5.299 |
| 2:00:00 PM | 1-1-3-5-2 | 52.13 | 0.305 | 2.196 |
| 3:00:00 PM | 3-5-4-1 | 61.01 | 0.198 | 0.000 |
| 4:00:00 PM | 3-5-4-1 | 60.94 | 0.210 | 0.000 |
| 5:00:00 PM | 3-5-4-1 | 61.20 | 0.153 | 0.000 |
| 6:00:00 PM | 3-5-4-1 | 60.95 | 0.191 | 0.000 |
| 7:00:00 PM -11:00:00 PM | 1-1-2-2-3-3 | 43.81 | 0.282 | 5.299 |

With the Congestion Avoidance objective, the optimal routes show a uniform pattern in AM and PM hours: from 10pm until 11am 1-4-5-3 is selected as the optimal route and from 12pm to 9pm 3-5-4-1 is selected as the optimal rush hour. During AM rush hours, TT Gap increases to 19.09, probably because the speed decreases from the free flow speed value. However, that the optimal routes lead to no extra travelled mileage (TD Gap is 0 for all time periods). Moreover, the objective function value during the PM peak hours, the busiest and most congested time of day in Miami, is still within an acceptable range (less than 0.15 increase in the total TTI). This indicates that congestion avoidance can be achieved without sacrificing the minimum distance. Nevertheless, this objective function increases the total travel time by 30%. This observation indicates that the concept of avoiding congestion may offer roadway users less congestion and higher speed levels, and consequently, faster routing is not necessarily an accurate assumption. Yet, this objective function can help minimizing the number of acceleration and deceleration movements, frustration of being stuck in traffic jams and consequently provides a more delightful travelling experience even though the duration is longer than usual. Crash risk minimization planning shows erratic changes at night. However, route 2-4-6-3 seems to be the safest route to take during both rush hours.

In the main objective function (safe and least congested routing), the impact of crash risk minimization is more dominant than TTI during midday (9am-2pm) and night (7pm-11pm) hours. This is because of the reduced congestion probability at those hours. During rush hours (7am-8am and 3pm-6pm), on the other hand, the routing is influenced by TTI minimization mostly. During the remaining times (12 am-5m), the two components of objective function have equal impact, and the solution is a trade-off between the two objectives. It can also be seen that safe routing increases travelled distance up to 5.6% (from Table 6 the maximum travelled distance gap for weighted objective function is 1.806; it is 5.6% of the minimum distance objective function of 32.30) and total travel time up to 31% (from Table 6 the maximum value of travel time gap for weighted sum objective function is 18.98; it is 31% of the minimum travel time of 61.20)

Whether these numbers justify the feasibility of this approach in a real world application or not, will mainly depend on policy makers and stakeholders. Type of business plays an important role in this decision making as well. For transporting vulnerable populations as well as hazardous material handling and medical substances among others, safety can be of higher priority than time and distance. Safe path selection can be a critical addition for the navigation devices, as a substitute for conventional minimum travel time routes.

Table 6. Case Study Routing And Scheduling Results. OFV: Objective Function Value. TT Gap: Travel Time Increase Compared to Routing with Minimum Travel Time, CR Gap: Increase in Crash Risk Compared to Routing with Minimum Crash Probability, TD Gap: Extra Travelled Mileage Compared to Routing with Minimum Travelled Distance. Routes are numbered according to arcs. Computational Time for all instance is less than 200 milliseconds for all scenarios.

| | | Weighted Sum of Objectives | | | | | Crash Risk Minimization | | | | Congestion Avoidance (TTI Delay Minimization) | | | | |
|---|---|---|---|---|---|---|---|---|---|---|---|---|---|---|---|
| | Start Time | Route | OFV | TT Gap | CR Gap | TD Gap | Route | OFV | TT Gap | TD Gap | Route | OFV | TT Gap | CR Gap | TD Gap |
| 1 | 12:00:00 AM | 1-6-5-2 | 1.575 | 11.63 | 0.000 | 0.900 | 1-6-5-2 | 0.608 | 11.63 | 0.900 | 1-4-5-3 | 1.00 | 19.09 | 0.038 | 0.00 |
| 2 | 1:00:00 AM | 2-5-6-1 | 1.542 | 11.63 | 0.014 | 0.900 | 2-2-3-6-1 | 0.561 | 3.50 | 4.002 | 1-4-5-3 | 1.00 | 19.09 | 0.089 | 0.00 |
| 3 | 2:00:00 AM | 1-6-5-2 | 1.470 | 11.63 | 0.012 | 0.900 | 1-1-3-5-2 | 0.491 | 8.14 | 2.196 | 1-4-5-3 | 1.00 | 19.09 | 0.062 | 0.00 |
| 4 | 3:00:00 AM | 2-5-6-1 | 1.509 | 11.63 | 0.087 | 0.900 | 1-1-2-2-3-3 | 0.455 | 0.00 | 5.299 | 1-4-5-3 | 1.00 | 19.09 | 0.130 | 0.00 |
| 5 | 4:00:00 AM | 2-5-6-1 | 1.548 | 11.63 | 0.000 | 0.900 | 2-5-6-1 | 0.580 | 11.63 | 0.900 | 1-4-5-3 | 1.00 | 19.09 | 0.057 | 0.00 |
| 6 | 5:00:00 AM | 2-5-6-1 | 1.529 | 11.63 | 0.022 | 0.900 | 1-1-2-5-3 | 0.540 | 8.14 | 2.196 | 1-4-5-3 | 1.00 | 19.09 | 0.063 | 0.00 |
| 7 | 6:00:00 AM | 2-4-6-3 | 1.590 | 13.24 | 0.000 | 1.806 | 2-4-6-3 | 0.612 | 13.24 | 1.806 | 1-4-5-3 | 1.00 | 17.62 | 0.106 | 0.00 |
| 8 | 7:00:00 AM | 1-4-5-3 | 1.719 | 13.72 | 0.000 | 0.000 | 1-4-5-3 | 0.739 | 13.72 | 0.000 | 1-4-5-3 | 1.05 | 13.72 | 0.000 | 0.00 |
| 9 | 8:00:00 AM | 1-4-5-3 | 1.963 | 13.64 | 0.125 | 0.000 | 3-6-4-2 | 0.843 | 25.33 | 1.806 | 1-4-5-3 | 1.12 | 13.64 | 0.125 | 0.00 |
| 10 | 9:00:00 AM | 2-4-6-3 | 1.741 | 11.93 | 0.000 | 1.806 | 2-4-6-3 | 0.720 | 11.93 | 1.806 | 1-4-5-3 | 1.10 | 15.85 | 0.079 | 0.00 |
| 11 | 10:00:00 AM | 2-4-6-3 | 1.709 | 12.21 | 0.000 | 1.806 | 2-4-6-3 | 0.693 | 12.21 | 1.806 | 1-4-5-3 | 1.10 | 16.28 | 0.065 | 0.00 |
| 12 | 11:00:00 AM | 3-6-4-2 | 1.677 | 12.15 | 0.000 | 1.806 | 3-6-4-2 | 0.654 | 12.15 | 1.806 | 1-4-5-3 | 1.12 | 16.23 | 0.184 | 0.00 |
| 13 | 12:00:00 PM | 2-4-6-3 | 1.690 | 11.78 | 0.000 | 1.806 | 2-4-6-3 | 0.663 | 11.78 | 1.806 | 3-5-4-1 | 1.13 | 15.86 | 0.155 | 0.00 |
| 14 | 1:00:00 PM | 3-6-4-2 | 1.716 | 11.62 | 0.000 | 1.806 | 3-6-4-2 | 0.688 | 11.62 | 1.806 | 3-5-4-1 | 1.14 | 15.70 | 0.130 | 0.00 |
| 15 | 2:00:00 PM | 3-6-4-2 | 1.705 | 4.98 | 0.000 | 1.806 | 3-6-4-2 | 0.674 | 4.98 | 1.806 | 3-5-4-1 | 1.13 | 8.95 | 0.269 | 0.00 |
| 16 | 3:00:00 PM | 3-5-4-1 | 1.925 | 0.00 | 0.198 | 0.000 | 3-6-4-2 | 0.729 | 4.93 | 1.806 | 3-5-4-1 | 1.12 | 0.00 | 0.198 | 0.00 |
| 17 | 4:00:00 PM | 3-5-4-1 | 1.995 | 0.00 | 0.210 | 0.000 | 2-4-6-3 | 0.790 | 22.19 | 1.806 | 3-5-4-1 | 1.11 | 0.00 | 0.210 | 0.00 |
| 18 | 5:00:00 PM | 3-5-4-1 | 1.944 | 0.00 | 0.153 | 0.000 | 2-4-6-3 | 0.788 | 22.10 | 1.806 | 3-5-4-1 | 1.15 | 0.00 | 0.153 | 0.00 |
| 19 | 6:00:00 PM | 3-5-4-1 | 1.947 | 0.00 | 0.191 | 0.000 | 2-4-6-3 | 0.762 | 12.53 | 1.806 | 3-5-4-1 | 1.12 | 0.00 | 0.191 | 0.00 |
| 20 | 7:00:00 PM | 2-4-6-3 | 1.665 | 12.68 | 0.000 | 1.806 | 2-4-6-3 | 0.659 | 12.68 | 1.806 | 3-5-4-1 | 1.05 | 16.69 | 0.228 | 0.00 |
| 21 | 8:00:00 PM | 3-6-4-2 | 1.590 | 13.66 | 0.000 | 1.806 | 3-6-4-2 | 0.611 | 13.66 | 1.806 | 3-5-4-1 | 1.01 | 18.08 | 0.070 | 0.00 |
| 22 | 9:00:00 PM | 3-5-4-1 | 1.557 | 18.67 | 0.002 | 0.000 | 2-5-6-1 | 0.587 | 11.40 | 0.900 | 3-5-4-1 | 1.00 | 18.67 | 0.002 | 0.00 |
| 23 | 10:00:00 PM | 1-4-5-3 | 1.603 | 18.98 | 0.000 | 0.000 | 1-4-5-3 | 0.636 | 18.98 | 0.000 | 1-4-5-3 | 1.00 | 18.98 | 0.000 | 0.00 |
| 24 | 11:00:00 PM | 1-6-5-2 | 1.595 | 11.63 | 0.000 | 0.900 | 1-6-5-2 | 0.627 | 11.63 | 0.900 | 1-4-5-3 | 1.00 | 19.09 | 0.058 | 0.00 |

## 6. Conclusion

We propose a two phase time dependent safe vehicle routing and scheduling optimization model that identifies the safest and least congested routes, as a substitute for the traditional objectives given in the literature such as shortest distance or travel time, through (1) avoiding recurring congestions, and (2) selecting routes that have a lower probability of crash occurrences and non-recurring congestion caused by those crashes. The paper introduces the idea of using crash risk and travel time delay through crash probability and travel time index (TTI) in vehicle routing problem. In the first phase, our model identifies the routing of a fleet and sequence of nodes on the safest feasible paths. The second phase reschedules the departure times of each vehicle from each node and adjusts the optimal speed on each arc. A modified simulated annealing (SA) algorithm is formulated to solve the NP-hard optimization problem. Results show that SA is capable of solving complex models iteratively in a considerably short computational time. We also introduce a new approach for estimations the speeds on arcs without real-life speed data via a queuing model. We apply the proposed models on a small real-world case study application in the City of Miami. Routing schema with respect to several conditions are evaluated for this case study application: (1) minimize the traffic delay (maximum congestion avoidance), (2) minimize traffic crash risk and (3) the combination of two. Using these results, we compare the travel times and traversed distances obtained from the proposed models to the classic objectives of travel time and distance minimization.

The results showed that safe routing can be achieved with none or slight increase in travelled distance. The effect of crash risk minimization outweighs TTI at night and during midday hours providing a safer travel for roadway users. However, during the PM rush hours, the routing is influenced by TTI minimization mostly especially due to the lower level of service (LOS). In some cases, the optimal route is different from either routes, and is a safer route through a trade-off between both objectives. We suggest the proposed safe routing approach to policy makers and planners since safer routing can be achieved with the proposed models for a slight increase in travelled distance or travel time. Traffic accidents are costly, and route planning through safer routes can result in major savings for fleet owners as well as insurance companies. The knowledge obtained from this approach can successfully contribute to the development of more reliable safety-focused transportation plans, as the solution of the model points to specific situations where routes can be selected based on minimizing the crash risk rather than only the travel time, which will, in turn, help improving the safety for the roadway users. Some transportation activities such as hazardous material handling and emergency transportation among others, require a higher safety of transportation. The proposed concepts in this paper can be implemented on other disciplines of transportation and logistics such as path finding and network design. Formulating the models in a multi-objective format to study the safe paths can be an interesting future work. The actual performance of safe routing in GPS devices for navigation is also an interesting future study direction. In addition, if the majority of vehicles on a network take the proposed approach in this paper, rerouting traffic may shift these areas to other network segments. Therefore, researchers can study the safety rerouting in network level and in aggregate setting. In this paper historic crash data was used for the estimation of risk factor. One can use prediction techniques while accounting for unobserved heterogeneity (please see [45]) to estimate the risk factor in future. Finally, the concept of safety in routing can be studied using other approaches, such as user equilibrium modelling, dynamic programming, and dynamic traffic assignment among others.

**Acknowledgements**

This project was supported by United States Department of Transportation grant DTRT13-G-UTC42, and administered by the Center for Accessibility and Safety for an Aging Population (ASAP) at the Florida State University (FSU), Florida A&M University (FAMU), and University of North Florida (UNF). We also thank the Florida Department of Transportation and National Oceanic and Atmospheric Administration for providing the data. The opinions, results, and findings expressed in this manuscript are those of the authors and do not necessarily represent the views of the United States Department of Transportation, The Florida Department of Transportation, The National Oceanic and Atmospheric Administration, The Center for Accessibility and Safety for an Aging Population, the Florida State University, the Florida A&M University, or the University of North Florida.